\documentclass[10pt,twocolumn,letterpaper]{article}

\usepackage[pagenumbers]{cvpr} 

\definecolor{linkc}{rgb}{0, 0.44, 0.74}
\definecolor{eqc}{rgb}{1, 0, 0}
\usepackage{url}
\usepackage{tabularx}
\usepackage{booktabs}
\usepackage{graphicx}
\usepackage{caption}
\usepackage{subcaption}
\usepackage{array}
\usepackage{enumitem}
\usepackage{wrapfig}
\usepackage{booktabs}
\usepackage{subcaption}
\usepackage[table]{xcolor}
\usepackage{placeins}
\usepackage{xcolor}

\definecolor{color1}{HTML}{BC4B51}    
\definecolor{view}{HTML}{8CB369}      
\definecolor{texture}{HTML}{F4A259}   
\definecolor{color2}{HTML}{5B8E7D}    

\usepackage[many]{tcolorbox}    	
\usepackage{setspace}               
\usepackage{float}

\definecolor{main}{HTML}{626172}    
\definecolor{sub}{HTML}{F5F5F5}     

\newtcolorbox{boxF}{
    breakable,
    fontupper = \color{black},
    rounded corners,
    arc = 6pt,
    colback = sub,
    enhanced,
    boxrule = 1.5pt, 
    colframe = white, 
    borderline = {1.5pt}{0pt}{main, dashed} 
}

\newcolumntype{Y}{>{\centering\arraybackslash}X}

\usepackage{caption}

\makeatletter
\renewcommand\paragraph{\@startsection{paragraph}{4}{\z@}%
  {0.4ex \@plus 0.15ex \@minus 0.1ex}
  {-0.7em}
  {\normalfont\normalsize\bfseries}}
\makeatother

\definecolor{cvprblue}{rgb}{0.21,0.49,0.74}
\usepackage[pagebackref,breaklinks,colorlinks,allcolors=cvprblue]{hyperref}

\title{Asymmetric Idiosyncrasies in Multimodal Models}

\author{
Muzi Tao$^{1}$\quad
Chufan Shi$^{1}$ \quad
Huijuan Wang$^{1}$ \quad
Shengbang Tong$^{2}$ \quad
Xuezhe Ma$^{1}$ \\[0.3em]
$^{1}$University of Southern California \\
$^{2}$New York University \\[0.2em]
{\tt\small \{muzitao, xuezhema\}@usc.edu}
}

\begin{document}
\maketitle
\begin{abstract}
In this work, we study idiosyncrasies in the caption models and their downstream impact on text-to-image models. We design a systematic analysis: given either a generated caption or the corresponding image, we train neural networks to predict the originating caption model. Our results show that text classification yields very high accuracy (99.70\%), indicating that captioning models embed distinctive stylistic signatures. In contrast, these signatures largely disappear in the generated images, with classification accuracy dropping to at most 50\% even for the state-of-the-art Flux model. To better understand this cross-modal discrepancy, we further analyze the data and find that the generated images fail to preserve key variations present in captions, such as differences in the level of detail, emphasis on color and texture, and the distribution of objects within a scene. Overall, our classification-based framework provides a novel methodology for quantifying both the stylistic idiosyncrasies of caption models and the prompt-following ability of text-to-image systems. Our project page is publicly available at \href{https://muzi-tao.github.io/asymmetric-idiosyncrasies/}{muzi-tao.github.io/asymmetric-idiosyncrasies}.
\end{abstract}
    
\section{Introduction}
\label{sec1}
Synthetic data now plays a central role in training and scaling multimodal systems \citep{brack2025train,hammoud2024synthclip,lai2024revisit}. In state-of-the-art image generation pipelines (e.g., DALL·E 3 \citep{DALLE3}, Playground v3~\citep{liu2024playground}, Qwen-image~\citep{wu2025qwen}), model-generated captions are used to expand training corpora and to refine text–image alignment. This practice implicitly assumes that captions are (i) stylistically neutral or at least interchangeable across captioning models, and (ii) faithfully convertible into visual content by text-to-image (T2I) models. Both assumptions are under-examined.

A growing body of work shows that language models imprint stable, model-specific fingerprints that enable source attribution from text \citep{geva2021transformer,mcgovern2025your,wanli2025imf,sun2025idiosyncrasies}. Similar dataset and model signatures have been reported in the vision domain \citep{torralba2011unbiased,corvi2023detection,you2025images, liu2025datasetbias}. However, it remains unclear whether \emph{caption-level} idiosyncrasies produced by vision–language models (VLMs or MLLMs)~\citep{anthropic2024claude35, OpenAI2022ChatGPT, liu2023visual, Gemini} propagate into the \emph{images} produced by downstream T2I systems. If such cross-modal transfer is weak, synthetic-caption pipelines could quietly introduce distributional biases at the text stage that do not materialize visually, complicating the use of captions as faithful supervisory signals.

We investigate this question using a simple, model-agnostic approach based on ``name-that-model'' classification on both sides of the caption–image interface. Given an image and a prompt, multiple captioning models produce captions; we first train a text classifier to attribute each caption to its source model. We then feed those same captions into a fixed T2I model and train an image classifier to attribute the \emph{generated images} to the caption source. If caption idiosyncrasies reliably transfer across modalities, attribution should remain high in the image domain; if not, we obtain a direct, quantitative, and interpretable measure of a cross-modal translation gap.

\begin{figure*}[htbp]
  \centering
  \includegraphics[width=0.99\textwidth]{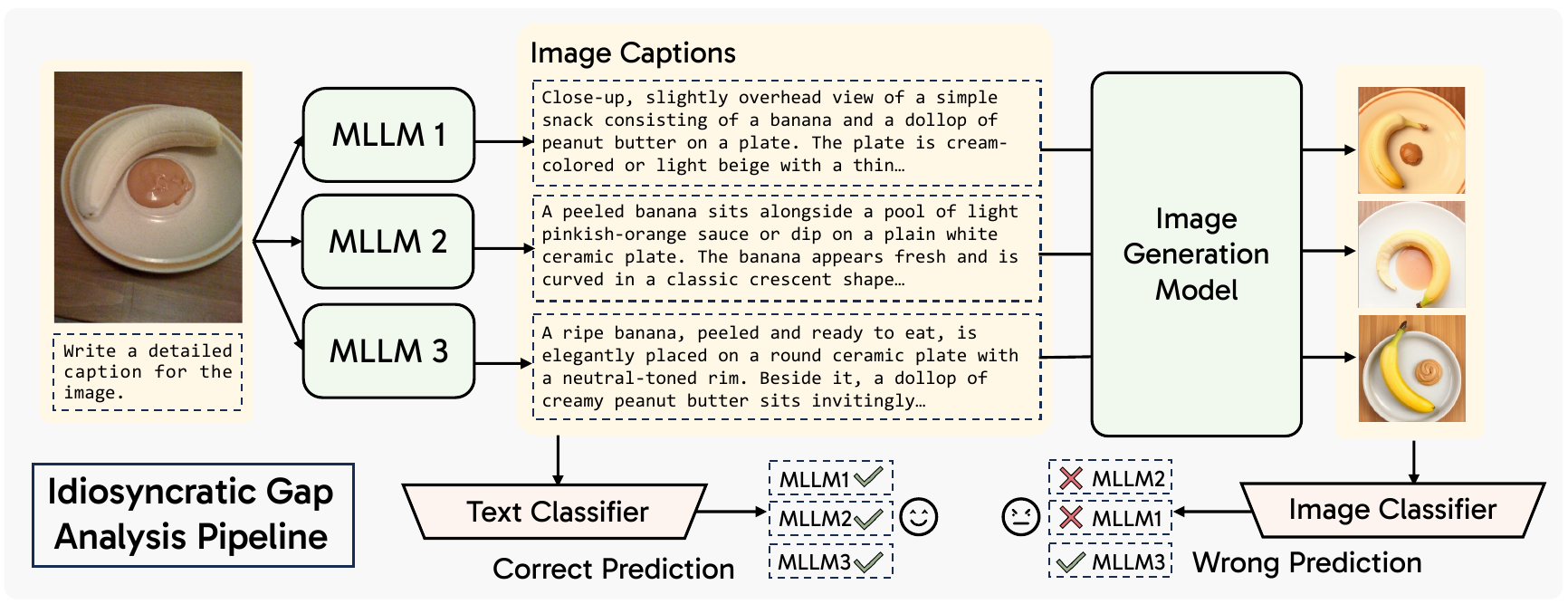}
  \caption{Overview of our pipeline. Multiple MLLMs generate captions for the same image, and a text classifier reliably attributes each caption to its source model. These captions are then used to synthesize new images, on which an image classifier performs the same source-identification task. While text-based attribution is highly accurate, image-based attribution fails, which reveals a clear mismatch between caption-space and image-space model signatures.}
  \label{fig:pipeline}
\end{figure*}

Empirically, caption idiosyncrasies are highly pronounced in text but largely dissipate when transferred into images. On 30k captions per model spanning diverse image sets, a straightforward BERT-based classifier~\citep{devlin2019bert} achieves 99.70\% accuracy in identifying the captioning model. In contrast, after rendering those captions with modern T2I systems, image-space attribution drops substantially, peaking at only around 50\% with the current best model Flux-schnell~\citep{labs2025flux1kontextflowmatching}, which is only modestly above the 33.3\% chance level for three classes. As a strong reference point, the identical image classifier achieves approximately 76.7\% accuracy when distinguishing \emph{natural} image sources of comparable scale, underscoring that the difficulty is specific to generated images rather than to the classifier itself.

To understand the behavior, we analyze linguistic and content features of captions. TF–IDF phrase statistics, color/texture vocabularies, and compositional terminology reveal stable, model-specific preferences (e.g., viewpoint/angle wording, ambience/lighting emphasis, or concise compositional framing). We find that paraphrasing the caption still preserves over \textbf{95\%} attribution accuracy. This provides strong evidence that the fingerprints extend beyond surface phrasing to choices about \emph{what} to describe and \emph{how} to structure it. Yet many of these choices fail to manifest reliably in the generated images, especially along axes such as level of detail, nuanced color/texture, and object layout.

These results provide an operational measure of a cross-modal idiosyncratic gap: stylistic and content-selection signals that are strong in captions are not faithfully realized by current T2I models. Practically, this suggests that (i) aggregating captions from diverse captioners may inject text-domain biases that do not become visual supervision, and (ii) instruction-following in T2I remains a key bottleneck for transferring caption semantics beyond object keywords. 
\begin{itemize}[leftmargin=*, itemsep=1pt, topsep=2pt]
    \item We propose a simple, scalable attribution framework that quantifies model-specific idiosyncrasies on both captions and the images generated from them.
    \item Using this framework, caption-based source attribution is nearly perfect at 99.70\%, whereas attribution from the corresponding generated images is substantially weaker, reaching only about 50\%. This contrast highlights a significant gap in cross-modal translation.
    \item Through lexical and structural analyses, including TF–IDF phrases, color and texture vocabularies, composition-related terms, and paraphrasing robustness tests, caption fingerprints are traced to deeper content-selection and perspective patterns. These patterns are not preserved by current T2I models.
    \item Finally, we propose attribution-as-evaluation as a complementary metric for prompt-following: instruction following should increase the transfer of caption idiosyncrasies into images, narrowing the gap.
\end{itemize}

\section{Background}
\label{sec2}
\subsection{Idiosyncrasies in Large Language Models}

Large language models achieve remarkable performance across diverse tasks by leveraging the statistical and semantic regularities embedded in large-scale corpora. Beyond their generalization capabilities, recent studies reveal that LLMs also exhibit stable, model-specific \emph{idiosyncrasies} in generated text. These idiosyncrasies, expressed as consistent stylistic and distributional patterns, act as implicit fingerprints that make outputs attributable to their source models \citep{geva2021transformer, mcgovern2025your, wanli2025imf}. Building on this view, \citet{sun2025idiosyncrasies} formalize an attribution task in which a classifier predicts the source model from generated samples. They show that such fingerprints persist across model families and prompting conditions, suggesting that differences extend beyond surface token statistics. Similar findings appear in authorship attribution and neural text forensics \citep{uchendu2020authorship, antoun2023source, dunlap2025vibecheck}, where stylistic or distributional features reveal model origin even under paraphrasing or translation. These observations naturally raise the question of whether analogous signatures also arise in multimodal settings.  

\subsection{Idiosyncrasies in Vision and Vision-Language Models}
Idiosyncratic signatures are not confined to text. In computer vision, classic studies show that simple classifiers can reliably distinguish between datasets, revealing systematic biases beyond semantic content \citep{torralba2011unbiased, corvi2023detection}. Similar effects are observed in generative models: diffusion- and GAN-based systems often imprint dataset- or model-specific artifacts that enable reliable attribution \citep{you2025images, mansour2024bias, tang2024enhancing}. These results suggest that high-dimensional visual data still carries persistent distributional cues.

With the rise of vision-language models (VLMs), such concerns extend across modalities. Models like CLIP learn joint embeddings of text and vision \citep{radford2021clip}, yet representations or downstream outputs may still reflect stylistic biases inherited from training. Recent work highlights that VLMs, when used for captioning or generation, can produce model-specific vocabulary, style, or narrative emphasis \citep{sun2025idiosyncrasies, dunlap2025vibecheck}. What remains unclear is whether these linguistic fingerprints propagate across modalities, specifically from captions into the images synthesized by downstream text-to-image systems. This question motivates our analysis.
\section{Idiosyncrasies in Generated Image Captions}
\label{sec3}

Prior research has demonstrated that large language models exhibit model-specific idiosyncrasies in their outputs. In this work, we ask: \textit{Does this observation apply to MLLMs and their downstream application, such as captioning?}

\subsection{Experimental Setup}

To investigate caption idiosyncrasies across different MLLMs, we formulate a classification task. Given a prompt \(p \in \mathcal{P}\) and an image \(x \in \mathcal{X}\), each model \(M_k\) produces a caption \(c = M_k(p,x)\), where \(\mathcal{C}_k\) is the set of captions from \(M_k\). Each caption \(c_i \in \mathcal{C}_k\) is paired with a label \(y_i = k\), indicating its source model. For \(K\) MLLMs, a \(K\)-way classifier is trained to predict \(y_i\) from \(c_i\). If caption distributions overlap heavily, accuracy should approach random guessing (\(1/K\)); substantially higher accuracy indicates model-specific linguistic fingerprints.

The image pool is constructed from several widely used datasets. Specifically, we sample 10,000 images in total: 3,000 each from the validation sets of CC3M~\citep{sharma2018conceptual}, COCO~\citep{lin2014microsoft}, and ImageNet~\citep{deng2009imagenet}, plus 1,000 from MNIST~\citep{deng2012mnist}.

For caption generation, we employ three proprietary MLLMs: Claude-3.5-Sonnet~\citep{anthropic2024claude35}, Gemini-1.5-Pro~\citep{team2024gemini}, GPT-4o~\citep{hurst2024gpt}, all accessed via their official APIs, and one open-sourced MLLM Qwen3-VL~\citep{qwen3technicalreport}. To capture linguistic diversity and range, we design three progressively detailed prompts for every image. These prompts elicit different levels of granularity and complexity, enabling a systematic comparison of captioning styles, lexical choices, and narrative depth across models under uniform prompting conditions. Specifically, the three prompts are as follows: 

\vspace{0.5em}
\noindent\textbf{Coarse captioning prompt: }
\begin{boxF}
Describe the image.
\end{boxF}
\noindent\textbf{Detailed captioning prompt: }
\begin{boxF}
Write a detailed caption for the image.
\end{boxF}
\noindent\textbf{Very detailed captioning prompt:}
\begin{boxF}
Tell me everything you can see in the image, including as many visible elements as possible.
\end{boxF}

The maximum output length is set to 1024 tokens for Prompts 1 and 2, and 4096 for Prompt 3, enabling more detailed descriptions. 
Captions were split 80\%/20\% randomly, yielding 72k training and 18k test samples, with all prompts of the same image allocated consistently. For classification, \texttt{BERT-base-uncased} \citep{devlin2019bert} was fine-tuned with a [CLS]-based linear head to predict the generating model. Training used the AdamW optimizer with a learning rate of \(2\times10^{-5}\), weight decay \(0.01\), batch size 32, 3 epochs, and dropout \(p=0.1\), with a linear decay schedule.

\begin{table}[!ht]
\small
\centering
\caption{Caption classification accuracy (\%).}
\label{tab:cap-test-accuracy}
\resizebox{\linewidth}{!}{\begin{tabular}{ccccc}
\toprule
\textbf{Claude-3.5-Sonnet} & \textbf{Gemini-1.5-Pro} & \textbf{GPT-4o} & \textbf{Qwen3-VL} & \textbf{Total} \\
\midrule
99.83 & 99.78 & 99.67 & --    & 99.76 \\
99.92 & 99.90 & --    & 99.73 & 99.85 \\
99.85 & --    & 99.62 & 99.13 & 99.53 \\
--    & 99.73 & 99.80 & 99.53 & 99.60 \\
99.62 & 99.65 & 99.57 & 99.30 & 99.53 \\
\bottomrule
\end{tabular}}
\end{table}

\subsection{Model-Specific Fingerprints in Captions}
As shown in Table~\ref{tab:cap-test-accuracy}, the classifier achieves an overall accuracy of \textbf{99.53\%}, far above the random baseline of 25\%. This near-perfect performance indicates that captions from different MLLMs contain highly distinctive linguistic signals, despite being generated under identical image and prompt conditions. In other words, outputs from Claude-3.5-Sonnet, Gemini-1.5-Pro, GPT-4o, and Qwen3-VL exhibit consistent stylistic or lexical fingerprints that enable reliable attribution.  

Per-class accuracies are also uniformly high, each exceeding 99.13\%. This suggests that fingerprints are not confined to a single model but are shared across all four. Together, these results confirm that stylistic biases are a systematic property of MLLM captioning rather than an isolated artifact.

\subsection{Word Distribution Analysis}

\begin{figure}[htbp]
    \centering
    \begin{subfigure}[b]{0.48\linewidth}
        \centering
        \includegraphics[width=\linewidth]{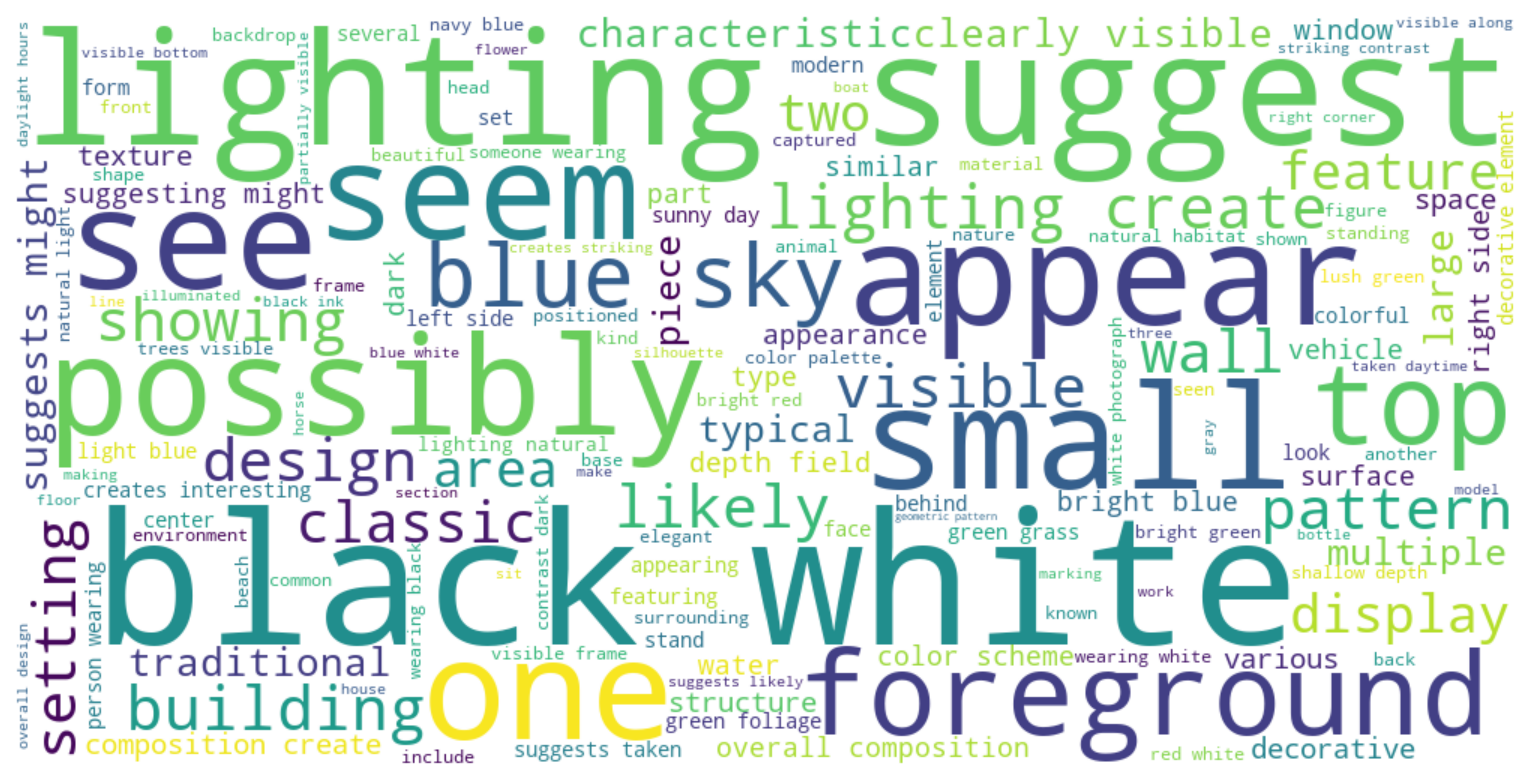}
        \caption{Claude-3.5-Sonnet}
    \end{subfigure}
    \hfill
    \begin{subfigure}[b]{0.48\linewidth}
        \centering
        \includegraphics[width=\linewidth]{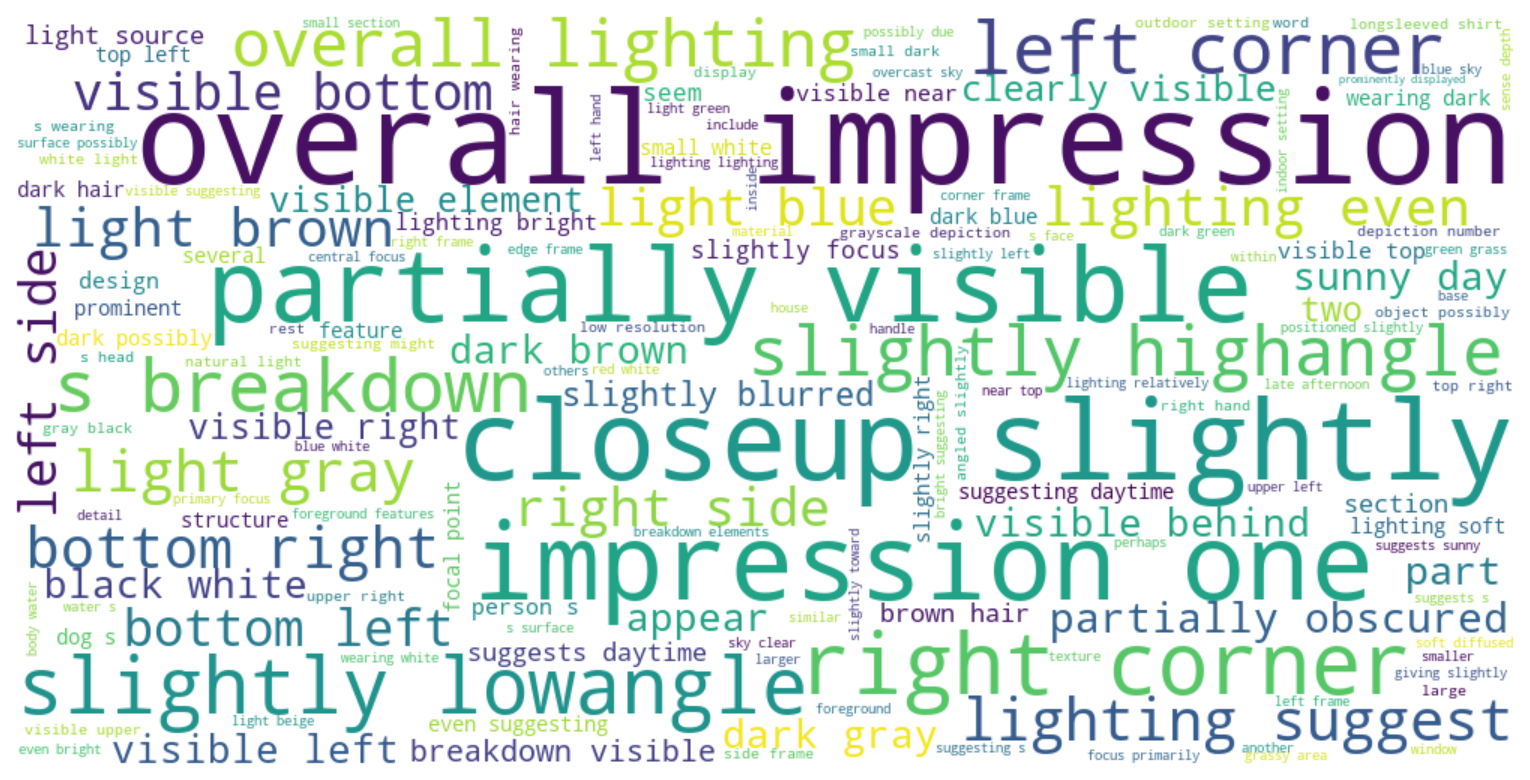}
        \caption{Gemini-1.5-Pro}
    \end{subfigure}
    
    \begin{subfigure}[b]{0.48\linewidth}
        \centering
        \includegraphics[width=\linewidth]{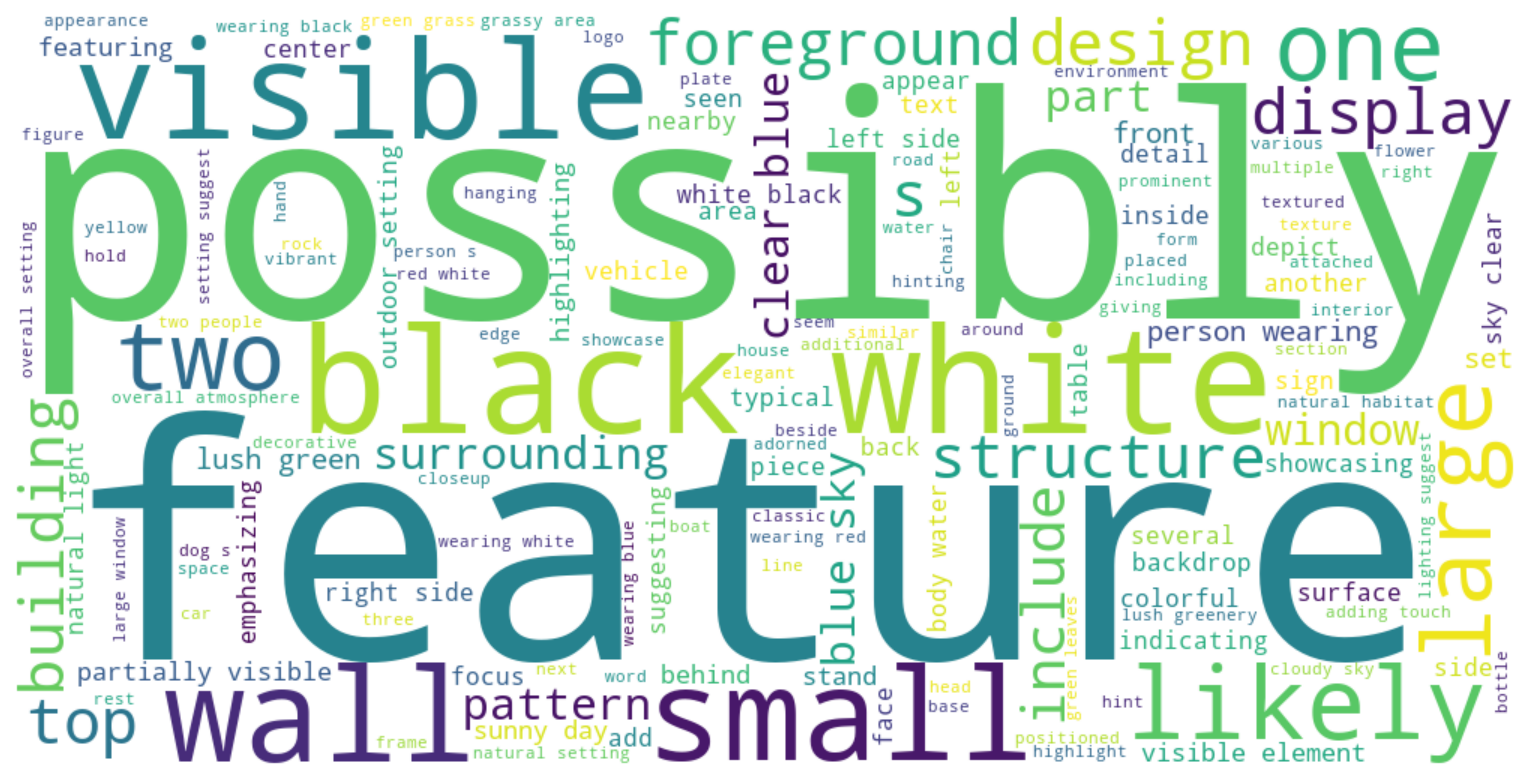}
        \caption{GPT-4o}
    \end{subfigure}
    \hfill
    \begin{subfigure}[b]{0.48\linewidth}
        \centering
        \includegraphics[width=\linewidth]{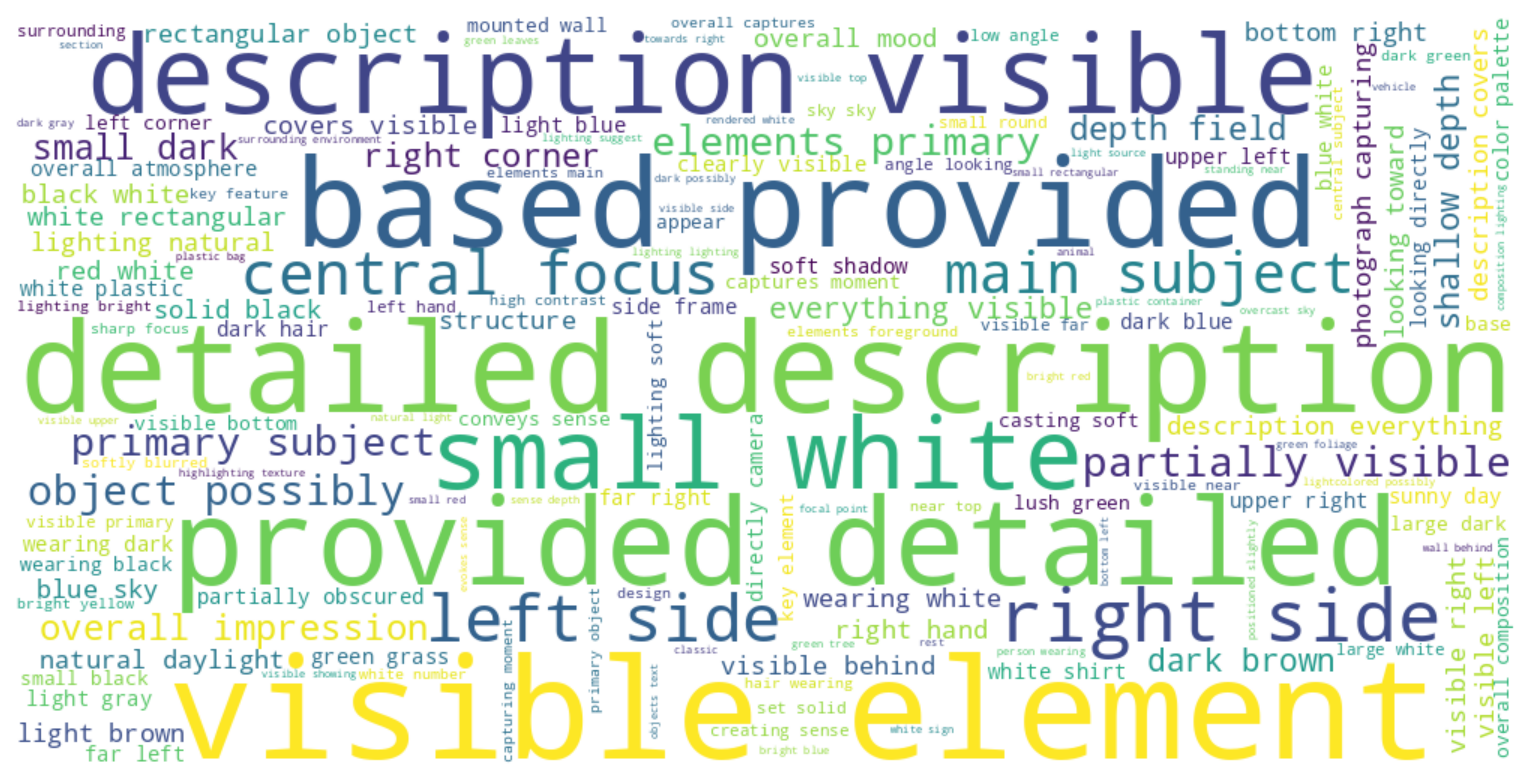}
        \caption{Qwen3-VL}
    \end{subfigure}
    \caption{Word clouds of model-generated captions}
    \label{fig:wordcloud}
\end{figure}

We next analyze word distributions to understand what makes captions from different models so separable. We apply TF-IDF~\citep{sparck1972statistical} ranking of 2-grams and 3-grams to the generated captions. For each model, we compute the top ten scoring phrases and list them in Table~\ref{tab:tfidf-top-phrases}, which provides representative lexical patterns. The results reveal clear biases: Claude frequently emphasizes lighting and visibility (e.g., ``lighting suggests,'' ``black,'' ``white''), Gemini highlights perspective and resolution (e.g., ``slightly low angle,'' ``impression,'' ``partially visible''), GPT favors categorical or structural terms (e.g., ``image depicts,'' ``feature,'' ``wall''), and Qwen centers on subjective prominence, contrast, and depth cues (e.g., ``central focus,'' ``depth field,'' ``high contrast''). These tendencies reflect stable narrative preferences: Claude focuses on ambience, Gemini emphasizes viewpoint, GPT prioritizes compositional framing, and Qwen highlights subject prominence. Word clouds (Figure~\ref{fig:wordcloud}) further confirm these stylistic differences.

\begin{table*}[ht!]
\centering
\small
\caption{Top distinctive TF-IDF phrases for each captioning model (generic appearance descriptors removed).}
\begin{tabularx}{\textwidth}{c X X X X}
\toprule
\textbf{Rank} &
\textbf{Claude-3.5-Sonnet} &
\textbf{Gemini-1.5-Pro} &
\textbf{GPT-4o} &
\textbf{Qwen3-VL} \\
\midrule
1  & lighting suggests            & overall impression      & image depicts            & detailed description \\
2  & visible background           & low resolution          & image features           & high contrast \\
3  & light colored                & slightly low angle      & person wearing           & central focus \\
4  & appears taken                & close slightly low      & handwritten digit        & main subject \\
5  & photo taken                  & eye level view          & lush green               & depth field \\
6  & lighting creates             & high angle view         & sunny day                & shallow depth field \\
7  & depth field                  & light gray              & clear blue               & black background digit \\
8  & composition creates          & partially visible       & handwritten number       & white number \\
9  & overall composition          & overall lighting        & setting appears          & visible elements \\
10 & scene appears                & slightly high angle     & partially visible        & image captures \\
\bottomrule
\end{tabularx}
\label{tab:tfidf-top-phrases}
\end{table*}

These findings show that the four models adopt stable yet distinct descriptive strategies. Such consistent preferences extend beyond individual words, forming recognizable linguistic fingerprints that explain the high classification accuracy. They further suggest that MLLMs inject narrative biases into captions, which may shape how downstream systems interpret the same images (Section~\ref{sec4}).

\section{Image Generation with Stylish Captions}
\label{sec4}

Given the strong idiosyncrasies observed in captions, a natural question is whether these stylistic signals transfer into the images generated from them. \textit{Do captions from different MLLMs yield visually distinctive images, or do generative models normalize such differences?} 

\subsection{Experimental Setup}

Our method adopts a parallel setup on the image side. Fixing a text-to-image generator \(G\), we use captions \(c_i \in \mathcal{C}_k\) from captioning models \(M_k \in \{M_1, M_2, \dots, M_K\}\) as input. This produces generated images
\[
\hat{x}_i = G(c_i), \quad \text{with label } y_i = k \text{ if } c_i \in \mathcal{C}_k .
\]

\begin{figure}
    \centering
    \includegraphics[width=0.9\linewidth]{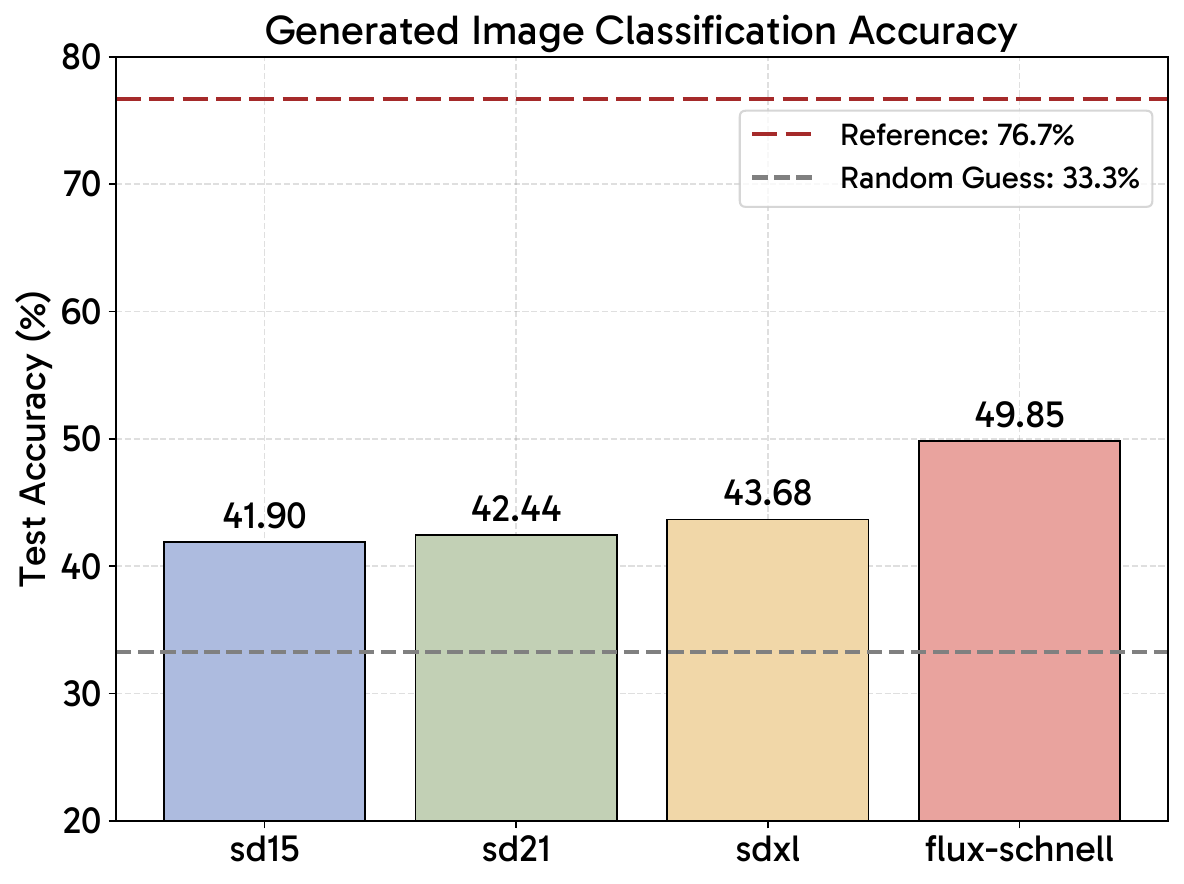}
    \caption{Classification performance on generated images. The test accuracy of image classification on generated images is 49.85\% with the SOTA model FLUX.1-schnell, while classification on a natural image dataset \citep{liu2025datasetbias} of the same scale using the same network achieves 76.7\%. Random guessing yields 33.3\%. }
    \label{fig:gen-img-acc}
\end{figure}

We then train an \(N\)-way classifier over the generated images \(\hat{x}_i\) to predict the originating captioning model \(M_k\). As in the text domain, classification accuracy above random chance (\(1/N\)) would indicate that model-specific idiosyncrasies persist in the generated images.

For image-side experiments, we use captions from Claude-3.5-Sonnet, Gemini-1.5-Pro, and GPT-4o, and render them with several widely used T2I systems: Stable Diffusion v1.5~\citep{rombach2022high}, Stable Diffusion v2.1~\citep{podell2023sdxl}, Stable Diffusion XL~\citep{podell2023sdxl}, and FLUX.1-schnell~\citep{esser2024scaling}. Following \citet{liu2025datasetbias}, we adopt comparable training settings for the image classifier. Specifically, we use a ResNet-18 backbone trained for 300 epochs with a batch size of 64, the AdamW optimizer (learning rate \(5\times10^{-4}\), weight decay 0.05), and standard augmentation including Mixup~\citep{zhang2017mixup} (\(\alpha=0.8\)) and CutMix~\citep{yun2019cutmix} (\(\alpha=1.0\)). We also apply label smoothing (0.1) and a warmup schedule of 20 epochs.

\subsection{Failure of Caption Fingerprints to Transfer}
Despite the near-perfect attribution observed for captions, classification on generated images is far less successful. As shown in Figure~\ref{fig:gen-img-acc}, the best-performing model (Flux-schnell) reaches only 49.85\% accuracy, barely above random guessing (33.3\%) and well below the 76.7\% accuracy achieved on natural images of similar scale \citep{liu2025datasetbias}. This indicates that the distinctive linguistic fingerprints of captions largely vanish once translated into the visual domain.  

\subsection{Ablation Study}

To better understand this discrepancy and further validate the finding, ablation studies are performed on top of the initial experiments.

\paragraph{Adding original images as a fourth class.}
To test whether the observed gap also holds against natural data, the 10k original images used as inputs to the captioning models are included as an additional class in the image classifier, yielding a 4-way classification setting with three generated-image classes and one natural-image class. The table below reports the results.

\begin{table}[H]
\vspace{-0.2em}
\centering
\small
\setlength{\tabcolsep}{4pt}
\begin{tabular}{lccccc}
\textbf{Model} & \textbf{Total} & \textbf{Claude} & \textbf{Gemini} & \textbf{GPT} & \textbf{Original} \\
\midrule
Accuracy (\%) & 51.84 & 50.83 & 56.31 & 38.30 & \cellcolor{green!10}82.11 \\
\end{tabular}
\vspace{-0.7em}
\end{table}

Overall accuracy in this setup is 51.84\%. Among the generated images, samples from Gemini are slightly more distinguishable; GPT is the hardest to classify. In contrast, the original images reach 82.1\% accuracy. This highlights a substantial gap in identifiable idiosyncrasy between natural and generated data.

\paragraph{Classification on keyword-prompts.}
To control for narrative style, we evaluate keyword-only prompts by first extracting keywords from each caption and then using the resulting keyword prompt to generate an image with FLUX-schnell. We run attribution on both the keyword prompt and the corresponding generated image. Results are below:

\begin{table}[H]
\vspace{-0.2em}
\centering
\small
\begin{tabular}{lcccc}
 & Gemini & GPT & Claude & Total \\
\midrule
keywords & 95.37 & 94.48 & 88.73 & \textbf{92.86} \\
images   & 46.67 & 41.34 & 41.56 & \textbf{43.22} \\
\end{tabular}
\vspace{-0.7em}
\end{table}

Text attribution remains high, while visual attribution stays low, indicating that captioner-specific cues are not primarily driven by stylistic fingerprints and are still largely not preserved through T2I generation.

\paragraph{Classification on extracted features.}
The dependence of the conclusion on the classifier architecture is further examined. Instead of training a convolutional network end-to-end, CLIP image features are extracted, and a linear classifier with SGD is trained with fixed hyperparameters across generation models. The test accuracies are reported below.

\begin{table}[H]
\vspace{-0.2em}
\centering
\small
\begin{tabular}{lcccc}
\textbf{Model} & \textbf{Flux-schnell} & \textbf{SDXL} & \textbf{SD 2.1} & \textbf{SD 1.5} \\
\midrule
Accuracy (\%) & 46.05 & 45.69 & 44.76 & 41.67 \\
\end{tabular}
\vspace{-0.7em}
\end{table}

Test accuracies remain low, ranging from 41.69\% to 46.05\%. The narrow spread and uniformly modest performance indicate that the difficulty in distinguishing generated images is not due to classifier choice or feature representation, but reflects the intrinsic similarity of the generated samples.

\section{The Idiosyncratic Gap Between Image Captioning and Generation Models}
\label{sec5}

The classification results raise a key question: \textit{Why are captions from different MLLMs easily distinguishable, while the corresponding generated images are not?}  

Intuitively, if the distinctive tokens in captions were faithfully mapped into the visual modality, their signatures should also appear in generated images. Moreover, given the vast pixel space and color range available, images should, in principle, be capable of encoding more information than a short caption. The failure of this transfer suggests that some informative features are lost during generation, whether genuinely valuable or merely stylistic.

\subsection{Linguistic Analysis on the Captions}

Attribution robustness to superficial cues is examined by modifying and paraphrasing captions before re-running classification. Simple edits include removing formatting, deleting special characters, and shuffling words or letters. In addition, we generate paraphrases using Qwen2.5~\citep{Yang2024Qwen25TR} with multiple prompting templates.

\begin{table*}[!t]
\centering
\small
\caption{Total and per-class accuracy under different text modifications and paraphrases. 
Paraphrasing was performed with three distinct prompts on \texttt{Qwen-2.5-1.5B-Instruct} and 
\texttt{Qwen-2.5-7B-Instruct} to ensure robustness across rewording styles and model scales. 
Detailed information about prompts is provided in Appendix~\ref{app:paraphrase_prompts}.}
\label{tab:robustness_accuracy}
\resizebox{0.8\textwidth}{!}{%
\begin{tabular}{lcccc}
\textbf{Text Transformation} & \textbf{Total} & \textbf{Claude-3.5-Sonnet} & \textbf{Gemini-1.5-Pro} & \textbf{GPT-4o} \\
\midrule
Removing Markdown Format & 99.71 & 99.73 & 99.62 & 99.77 \\
Removing Special Characters & 99.78 & 99.78 & 99.78 & 99.77 \\
Shuffling Words & 99.42 & 99.43 & 99.60 & 99.23 \\
Shuffling Letters & 34.49 & 0.00 & 100.00 & 3.48 \\
\midrule
Paraphrase 1 (\texttt{Qwen-2.5-1.5B-Instruct}) & 95.59 & 94.35 & 95.45 & 97.95 \\
Paraphrase 1 (\texttt{Qwen-2.5-7B-Instruct}) & 95.90 & 92.68 & 97.73 & 97.30 \\
Paraphrase 2 (\texttt{Qwen-2.5-1.5B-Instruct}) & 97.28 & 95.90 & 97.78 & 98.17 \\
Paraphrase 2 (\texttt{Qwen-2.5-7B-Instruct}) & 97.90 & 96.43 & 99.10 & 98.17 \\
Paraphrase 3 (\texttt{Qwen-2.5-1.5B-Instruct}) & 96.31 & 94.87 & 96.50 & 97.57 \\
Paraphrase 3 (\texttt{Qwen-2.5-7B-Instruct}) & 95.81 & 90.97 & 98.47 & 98.02 \\
\end{tabular}
}
\end{table*}

As shown in Table~\ref{tab:robustness_accuracy}, idiosyncrasies lie primarily at the word level rather than at individual characters, consistent with prior findings on LLMs~\citep{sun2025idiosyncrasies}. Even after paraphrasing, classification accuracy remains above 90\%, confirming that model-specific signals are not reducible to surface form, but reflect deeper factors such as descriptive perspective and content selection. Further qualitative analyses indicate that Claude-3.5-Sonnet tends to adopt a narrative, context-oriented framing, Gemini-1.5-Pro emphasizes camera perspective and exhaustive detail, and GPT-4o produces concise summaries focusing on salient objects and layout. These differences motivate a closer examination of how models encode and transmit visual content.

\subsection{Probing T2I Encoders}
The robustness to paraphrasing indicates that caption idiosyncrasies reflect deeper structural choices rather than superficial phrasing. This raises a natural question: does this signal survive the first stage of the text-to-image pipeline, the text encoder?

To examine this, we analyze the core encoders used in modern T2I systems. We reuse the linear-probe setup from our text classification task, but apply it directly to the final embeddings produced by T5~\citep{roberts2019exploring} and CLIP~\citep{radford2021learning}. If the encoder collapses or homogenizes stylistic cues, classification accuracy should drop substantially.

As shown in Table~\ref{tab:encoder_probe_results}, both encoders retain model-specific signals to a substantial degree. The T5 encoder preserves nearly all stylistic information (99.74\%), while CLIP also maintains high separability (94.14\%). These results indicate that the text encoders faithfully transmit caption-level fingerprints rather than discarding them.
This finding rules out the encoder as the primary source of the cross-modal gap. Since the stylistic signal reaches the generator largely intact, the discrepancy must arise in later stages of the pipeline. We therefore turn to the generation process itself to understand where the signal is diminished.

\begin{table}[h]
\centering
\small
\caption{Linear probe accuracy(\%) on T2I text encoder embeddings. The stylistic signal is overwhelmingly preserved, not lost.}
\label{tab:encoder_probe_results}
\begin{tabular}{lcccc}
\textbf{Encoder} & \textbf{Claude} & \textbf{Gemini} & \textbf{GPT} & \textbf{Average}\\
\midrule
CLIP & 94.47 & 95.1 & 92.85& 94.14\\
T5 & 99.73 & 99.82 & 99.67& 99.74 \\
\end{tabular}
\end{table}

\subsection{Image--Prompt Attribution in CLIP Space}

To test whether textual idiosyncrasies survive the generation process, we design a 3-way image–prompt attribution task. For each generated image, we retrieve the three captions with the same prompt index (one from each model). Since the image is produced from exactly one of them, the classifier must choose which captioning model is correct.

We freeze a CLIP text encoder, use precomputed image embeddings, and learn small linear projections that map both modalities into a shared space. If model-specific signals in the captions were faithfully transferred into the images, the classifier should recover the originating model with high accuracy.

\begin{table}[h]
\centering
\small
\begin{tabular}{lcccc}
\textbf{Model} & \textbf{Claude} & \textbf{Gemini} & \textbf{GPT} & \textbf{Average} \\
\midrule
Accuracy (\%) & 54.40 & 55.00 & 49.63 & 53.01 \\
\end{tabular}
\end{table}

As shown in the above table, attribution accuracy is only marginally above chance and far below the near-perfect scores achieved in the caption domain.  Even with the correct three candidate texts provided, the classifier cannot reliably recover which model produced the image, indicating that most textual signatures do not clearly manifest in the visual output.
This weak transfer motivates a closer examination of what visual content is actually preserved during generation, which we investigate next.

\subsection{Visual Content Analysis}
\label{visual_analysis}

\begin{figure*}[htbp]
    \centering
    \includegraphics[width=0.95\textwidth]{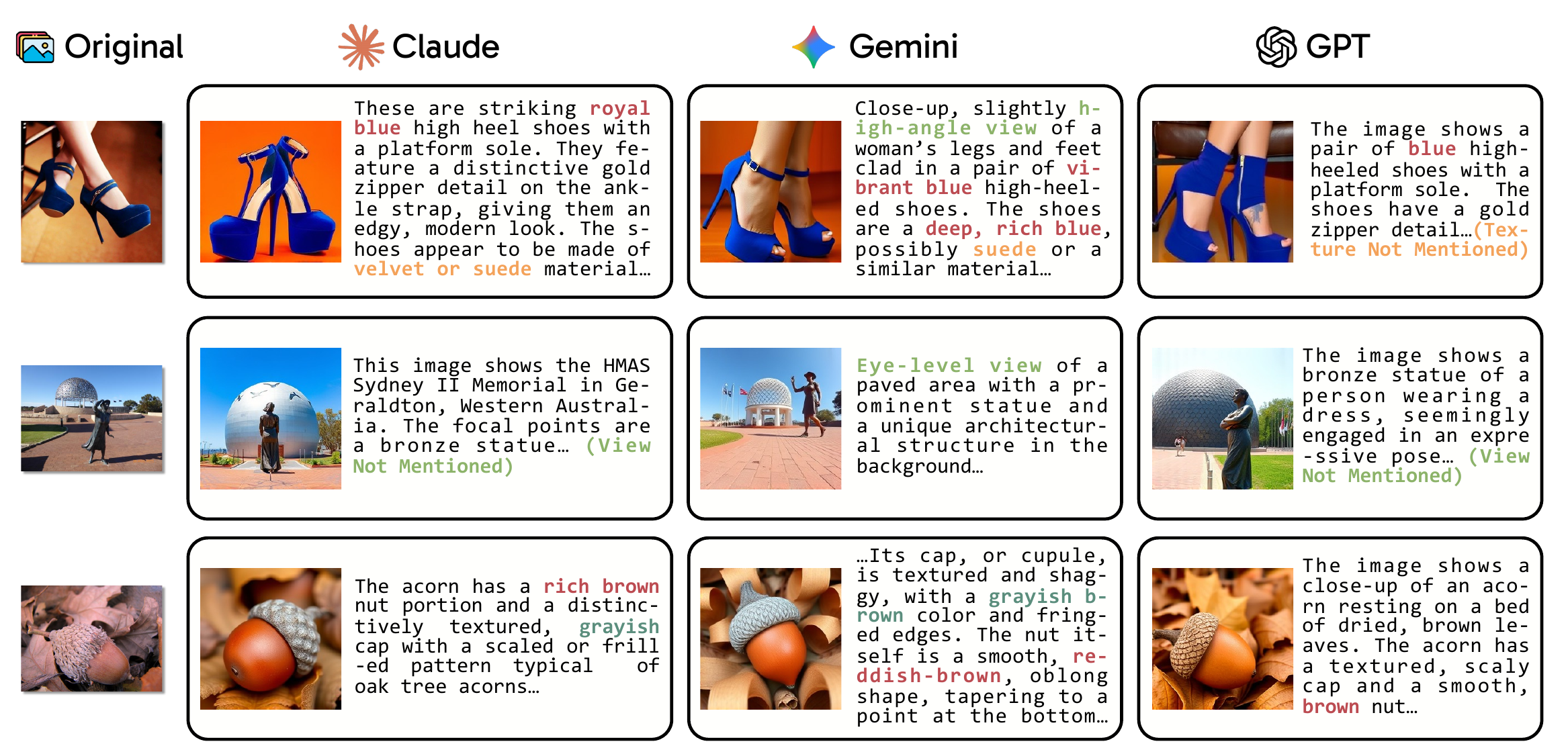}
    \caption{Comparison of captions generated by Claude-3.5-Sonnet, Gemini-1.5-Pro, and GPT-4o on the same images. Each row shows the original image on the left and, on the right, the captions generated by the three models and the images synthesized from those captions. The model outputs reveal several systematic failures across different attribute types. (i) In the first row, a simple descriptive \textcolor{color1}{color} term, such as blue, without any accompanying \textcolor{texture}{texture} specification, leads all models to produce images with broadly similar color–texture effects. However, none of the models reproduces the true, darker \textcolor{color1}{color} in the original image. (ii) In the second column, even though captions include explicit \textcolor{view}{view} descriptions, the generated viewpoints remain inconsistent: a caption describing a high-angle \textcolor{view}{view} yields an eye-level rendering, while an eye-level description produces a low-angle output. (iii) In the third row, different \textcolor{color1}{color} terms used to describe the acorn result in images that are still visually similar across models, and none of them match the actual color or appearance of the original image.}
    \label{fig:example}
\end{figure*}

The attribution experiments above show a consistent pattern: model-specific idiosyncrasies are strong and robust in text, yet much weaker in the generated images. To better understand where this information is attenuated, we complement the quantitative results with a more detailed analysis of the generated content.

Figure~\ref{fig:example} presents representative examples. For each real image, we show the three captions produced by Claude-3.5-Sonnet, Gemini-1.5-Pro, and GPT-4o, together with the images synthesized from those captions. Although the captions differ substantially in descriptive detail, color terminology, viewpoint specification, and texture vocabulary, the generated images remain visually similar across models. In many cases, even explicit attributes in the caption (e.g., ``dark blue'', ``shot from a high angle'', ``textured cap'') do not reliably appear in the output. This motivates a more systematic analysis along four dimensions.

\begin{figure*}[t]
    \centering
    \includegraphics[width=0.8\textwidth]{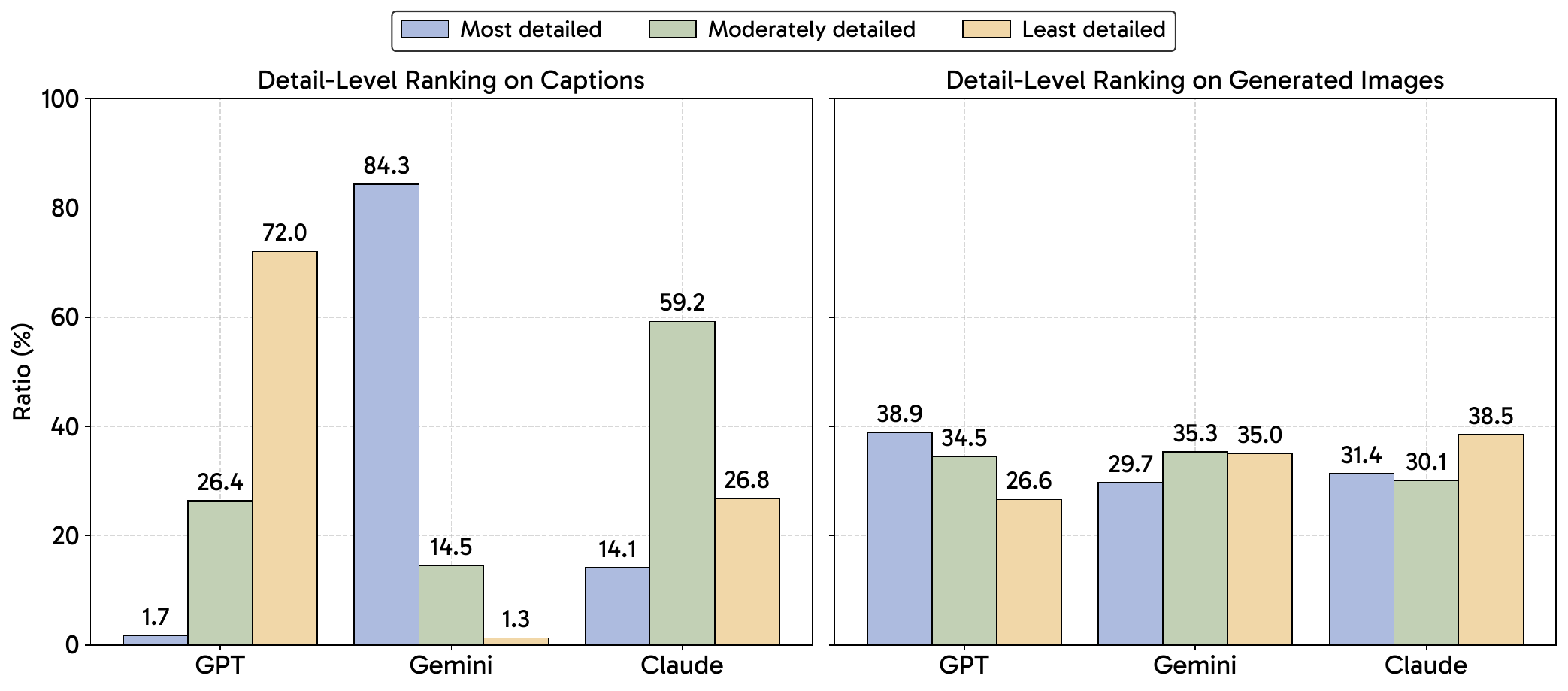}
    \caption{Detail-level rankings of both captions and generated images. Left: distribution of most-, moderately-, and least-detailed captions across the three models. Right: corresponding rankings assigned to the generated images.}
    \label{fig:detail_analysis}
\end{figure*}

\paragraph{Level of descriptive detail.}
Descriptive richness is evaluated using a ranking model that orders captions from most to least detailed, where \textit{detail} is defined as the amount of specific, factual, and descriptive information provided. Ranking is performed using \texttt{Qwen2.5-7B-Instruct}.
For each prompt, the three captions (from Claude-3.5-Sonnet, Gemini-1.5-Pro, and GPT-4o) are randomly shuffled and anonymized before evaluation to avoid bias. As shown in the left panel of Fig.~\ref{fig:detail_analysis}, Gemini-1.5-Pro stands out, with 84.27\% of its captions judged most detailed. GPT-4o is ranked last in 71.96\% of cases, while Claude-3.5-Sonnet most frequently occupies the middle rank (59.16\%). These results reveal a clear hierarchy of descriptive richness: \texttt{Gemini} $>$ \texttt{Claude} $>$ \texttt{GPT}.

\begin{table*}[htbp]
\centering
\caption{Lexical statistics of color and texture vocabulary in model-generated captions (30k per model). Metrics include total counts, percentage of captions containing at least one term, and average frequency per caption. Light green marks the highest value in each block, and light magenta marks the lowest.}
\label{tab:color_texture_stats}

\footnotesize  
\renewcommand{\arraystretch}{1.1} 

\resizebox{0.8\textwidth}{!}{%
\begin{tabular}{lcccccc}
\textbf{Model} & 
\shortstack{\textbf{Basic}\\\textbf{(Total)}} & 
\shortstack{\textbf{Nuanced}\\\textbf{(Total)}} & 
\shortstack{\textbf{With Basic}\\\textbf{(\%)}} &
\shortstack{\textbf{With Nuanced}\\\textbf{(\%)}} & 
\shortstack{\textbf{Basic}\\\textbf{(Average)}} & 
\shortstack{\textbf{Nuanced}\\\textbf{(Average)}}  \\
\midrule
\multicolumn{7}{c}{\textit{Color Vocabulary}} \\
Claude-3.5-Sonnet & 88,797  & 23,235  & 92.36 & 43.96 & 2.96 & 0.77 \\
Gemini-1.5-Pro    & \cellcolor{green!10}155,363 & \cellcolor{green!10}38,495  & \cellcolor{green!10}97.45 & \cellcolor{green!10}55.47 & \cellcolor{green!10}5.18 & \cellcolor{green!10}1.28 \\
GPT-4o            & \cellcolor{magenta!10}62,843  & \cellcolor{magenta!10}10,186  & \cellcolor{magenta!10}81.01 & \cellcolor{magenta!10}22.95 & \cellcolor{magenta!10}2.09 & \cellcolor{magenta!10}0.34 \\
\midrule
\multicolumn{7}{c}{\textit{Texture Vocabulary}} \\
Claude-3.5-Sonnet & \cellcolor{magenta!10}27,009 & 32,034 & \cellcolor{magenta!10}67.24 & 52.91 & \cellcolor{magenta!10}0.90 & 1.07 \\
Gemini-1.5-Pro    & \cellcolor{green!10}35,269 & \cellcolor{green!10}49,296 & \cellcolor{green!10}73.00 & \cellcolor{green!10}64.10 & \cellcolor{green!10}1.18 & \cellcolor{green!10}1.64 \\
GPT-4o            & 27,859 & \cellcolor{magenta!10}26,862 & 67.67 & \cellcolor{magenta!10}50.83 & 0.93 & \cellcolor{magenta!10}0.90 \\
\end{tabular}
}
\end{table*}

Similarly, \texttt{GPT-5} assigns detail-level rankings to generated images. The samples are again shuffled and anonymized to prevent bias. As shown in the right panel of Fig.~\ref{fig:detail_analysis}, however, the ordering is nearly reversed: the three models produce images with much more similar levels of detail, and images generated from GPT-4o captions are judged slightly more detailed overall. This indicates that the richness present in the captions is not faithfully retained during text-to-image generation.

\begin{figure}[t]
    \centering
    \includegraphics[width=0.8\linewidth]{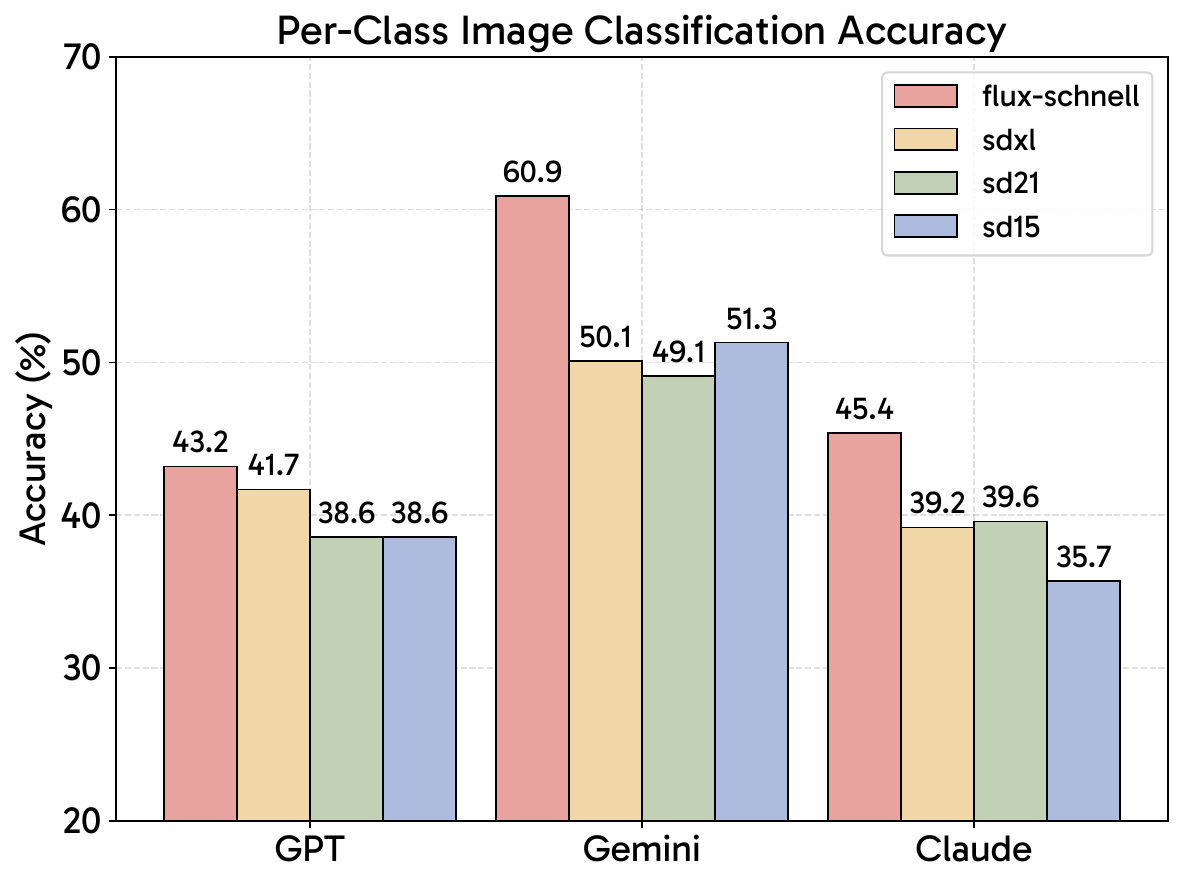}
    \caption{Per-class classification accuracy for generated images. Each bar shows how well images generated from captions of a given model can be attributed to their caption source.}
    \label{fig:image_cls_acc}
\end{figure}

\paragraph{Color vocabulary.}
The use of color terms is quantified with a deterministic dictionary-based matcher applied to normalized captions, counting both basic colors (e.g., red, green, blue) and nuanced variants (e.g., CSS/X11 shades, multi-word forms, shade modifiers). As shown in Table~\ref{tab:color_texture_stats}, Gemini-1.5-Pro shows the highest frequency and coverage of color terms; Claude-3.5-Sonnet exhibits similar usage but with a slightly broader nuanced vocabulary; and GPT-4o uses color terms least often with the narrowest set. Yet these pronounced textual gaps do not yield proportionate separability in the image domain (Fig.~\ref{fig:image_cls_acc}), suggesting that nuanced color instructions are often normalized by T2I models.

\begin{table}[h]
\centering
\scriptsize
\setlength{\tabcolsep}{8.5pt}
\caption{Semantic composition analysis of captions. Values indicate the percentage (\%) of captions meeting each criterion. }
\label{tab:semantic_composition_summary}
\begin{tabular}{lcccc}
    & \textbf{Spatial} & \textbf{Subject} & \textbf{Guiding} & \textbf{Balance} \\
\textbf{Model} 
    & \textbf{Layers} 
    & \textbf{Focus} 
    & \textbf{Elements} 
    & \textbf{Symmetry} \\
\midrule
Claude-3.5-Sonnet & \cellcolor{green!10}93.38 & \cellcolor{green!10}96.55 & \cellcolor{green!10}86.67 & \cellcolor{green!10}3.52 \\
Gemini-1.5-Pro    & 90.70 & 90.09 & \cellcolor{magenta!10}66.99 & \cellcolor{magenta!10}0.54 \\
GPT-4o            & \cellcolor{magenta!10}86.42 & \cellcolor{magenta!10}88.61 & 70.48 & 1.07 \\
\end{tabular}
\end{table}

\paragraph{Texture vocabulary.} We assess captions for the use of texture descriptors, distinguishing between basic tactile terms (e.g., rough, smooth) and more nuanced expressions for materials, finishes, or fine-grained surface qualities. Judgments are obtained using \texttt{Qwen2.5-7B-Instruct}. As shown in Table~\ref{tab:color_texture_stats}, Gemini-1.5-Pro employs texture vocabulary most extensively, particularly nuanced terms, while Claude-3.5-Sonnet and GPT-4o use fewer such descriptors. Overall, Gemini demonstrates the richest texture lexicon, Claude is moderate, and GPT is the most limited. Again, no matching ordering is observed in image attribution (Fig.~\ref{fig:image_cls_acc}), consistent with the hypothesis that fine-grained material cues are weakly realized by current T2I systems.

\paragraph{Visual composition.} A semantic analysis of captions is performed to assess whether they encode key principles of photographic composition. Using \texttt{Qwen2.5-1.5B-Instruct}, we evaluate each caption against four criteria: (1) explicit description of spatial layers (foreground, middle ground, background), (2) identification of a main subject and its focus state, (3) mention of guiding elements such as leading lines or framing, and (4) reference to balance, symmetry, or subject placement. 

Across 90,000 captions from 3 models, we observe clear differences in compositional awareness (Table~\ref{tab:semantic_composition_summary}). Claude-3.5-Sonnet consistently attains the highest coverage across all four criteria. In contrast, Gemini-1.5-Pro and GPT-4o score slightly lower on spatial layering and subject focus, and substantially lower on guiding elements and symmetry. Crucially, heightened compositional explicitness in text does not manifest as stronger per-class separability in images (Fig.~\ref{fig:image_cls_acc}), implying that composition-related instructions are partially lost or regularized by the generator.

\section{Discussion}
In this work, we present systematic evidence of a pronounced gap between image captioning and generation models. We show that the source model of a caption can be identified with near-perfect accuracy from text alone. However, these model-specific fingerprints largely vanish once captions are translated into images by current generators. Lexical, structural, and content analyses suggest that the gap stems not from surface phrasing, but from deeper descriptive choices that are inconsistently realized in images.

Our analysis shows that image generation models often fail to preserve fine-grained details in captions, contributing to the idiosyncratic gap between captioning and generation. As discussed in Section~\ref{visual_analysis}, the level of detail in captions is not fully reflected in the generated images. Nuanced color descriptions, for instance, rarely affect how colors are rendered. For composition, generation models may default to common scene structures based on context rather than strictly following specific words. These results highlight concrete limitations in current generation models and point to directions for future improvement.

\clearpage
{
    \small
    \bibliographystyle{ieeenat_fullname}
    \bibliography{main}
}
\clearpage
\appendix

\section{Human Baseline: User Study}
\label{app:user_study}

\begin{figure*}
    \centering
    \includegraphics[width=0.49\textwidth]{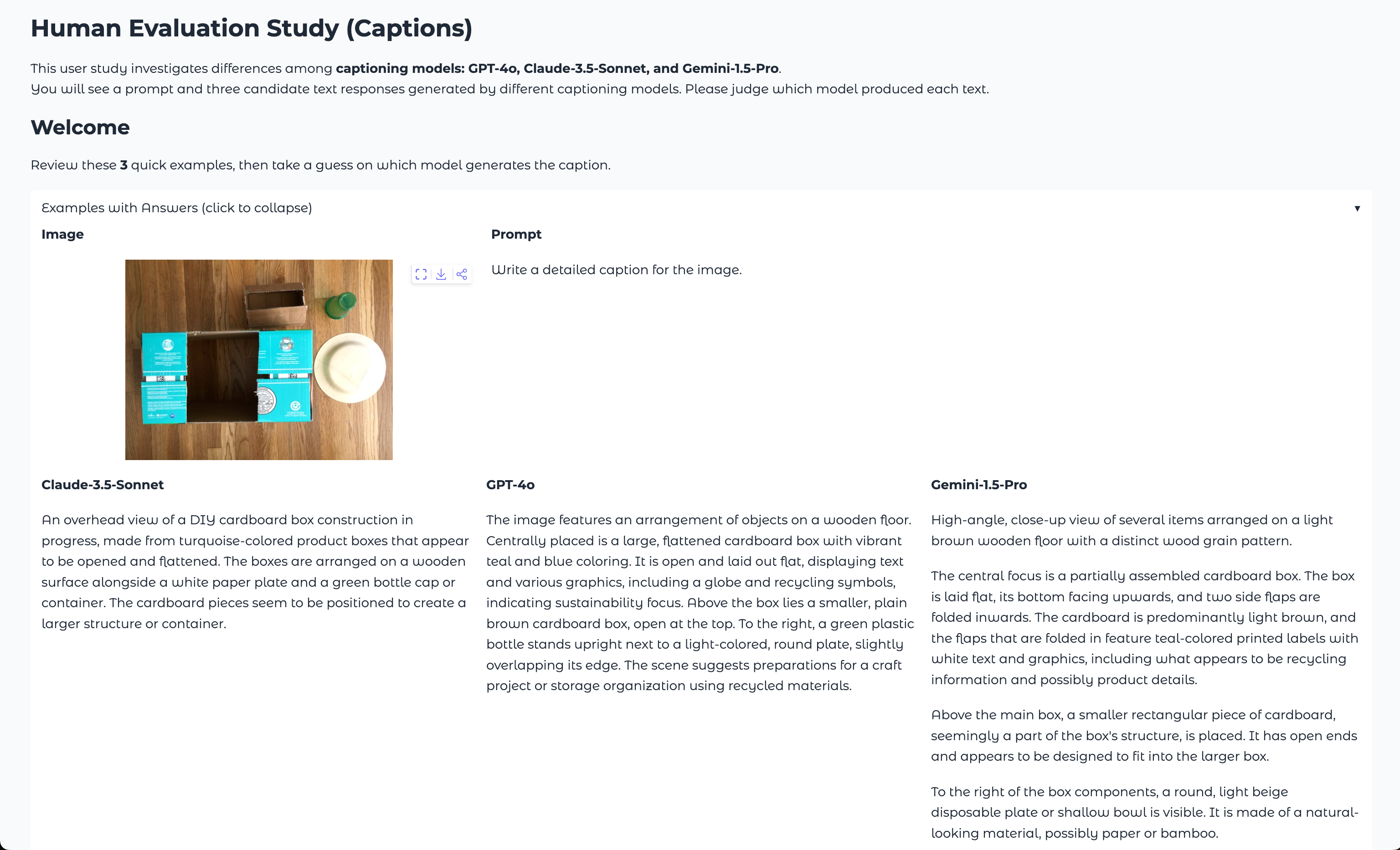}\hfill
    \includegraphics[width=0.49\textwidth]{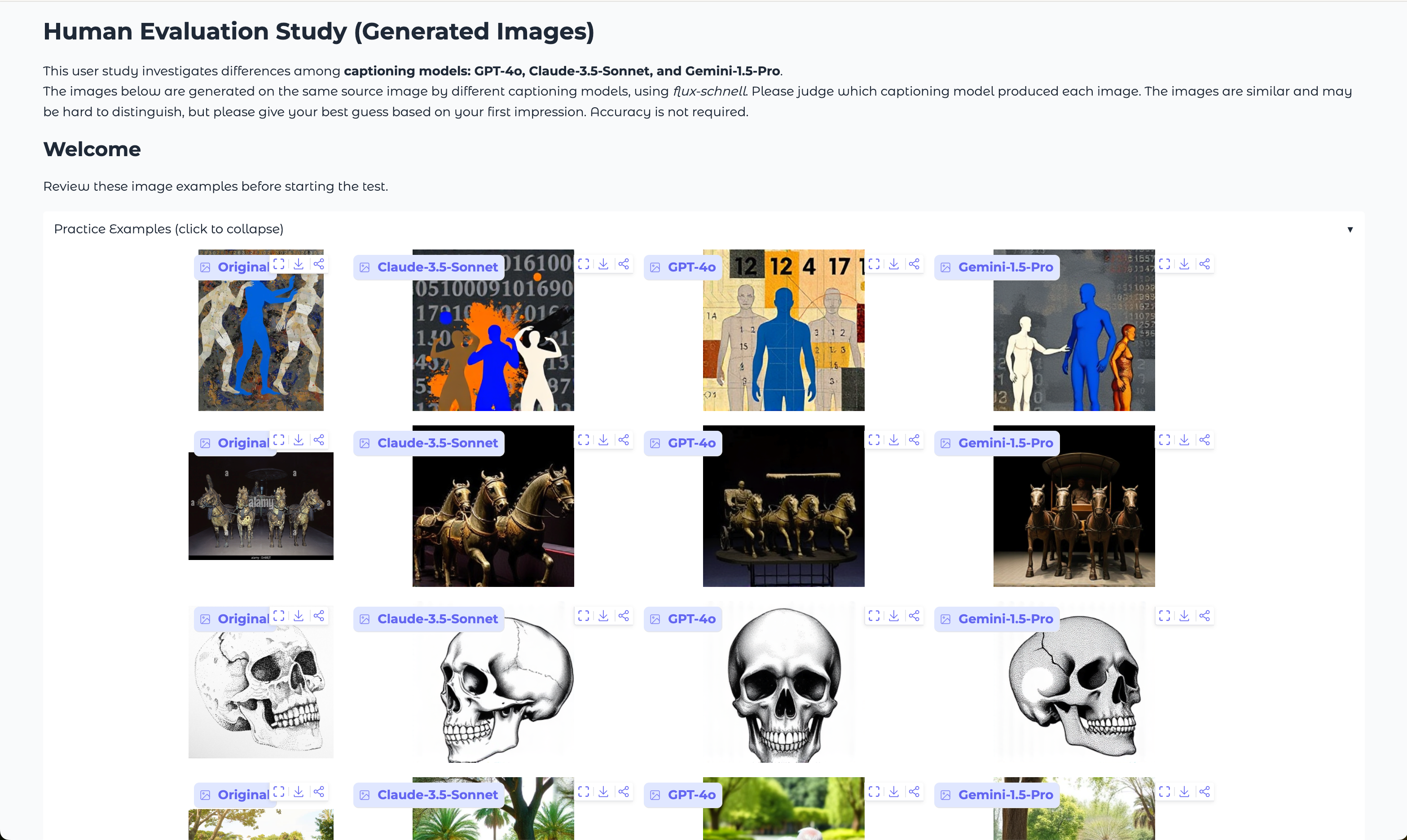}

    \vspace{6pt}

    \includegraphics[width=0.49\textwidth]{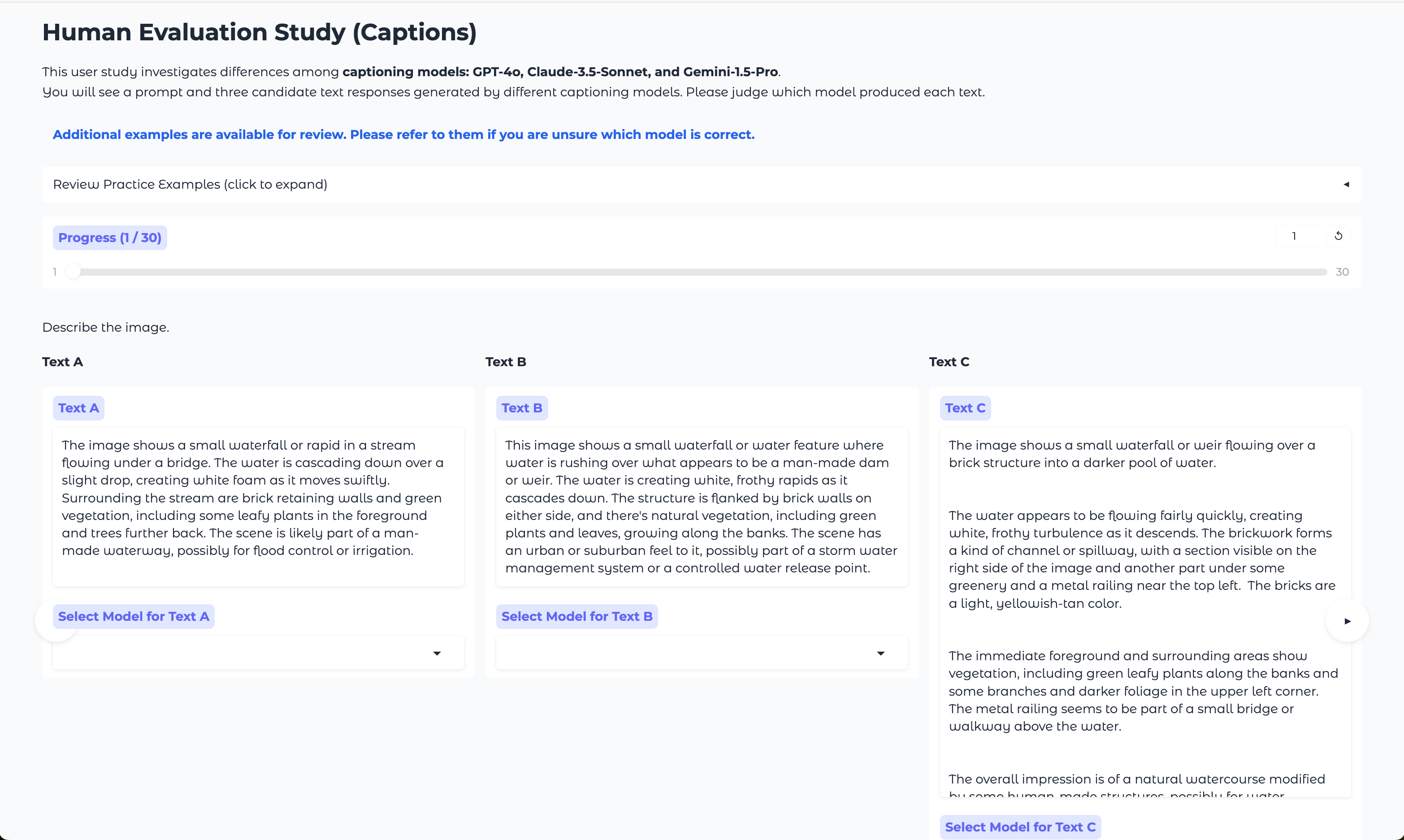}\hfill
    \includegraphics[width=0.49\textwidth]{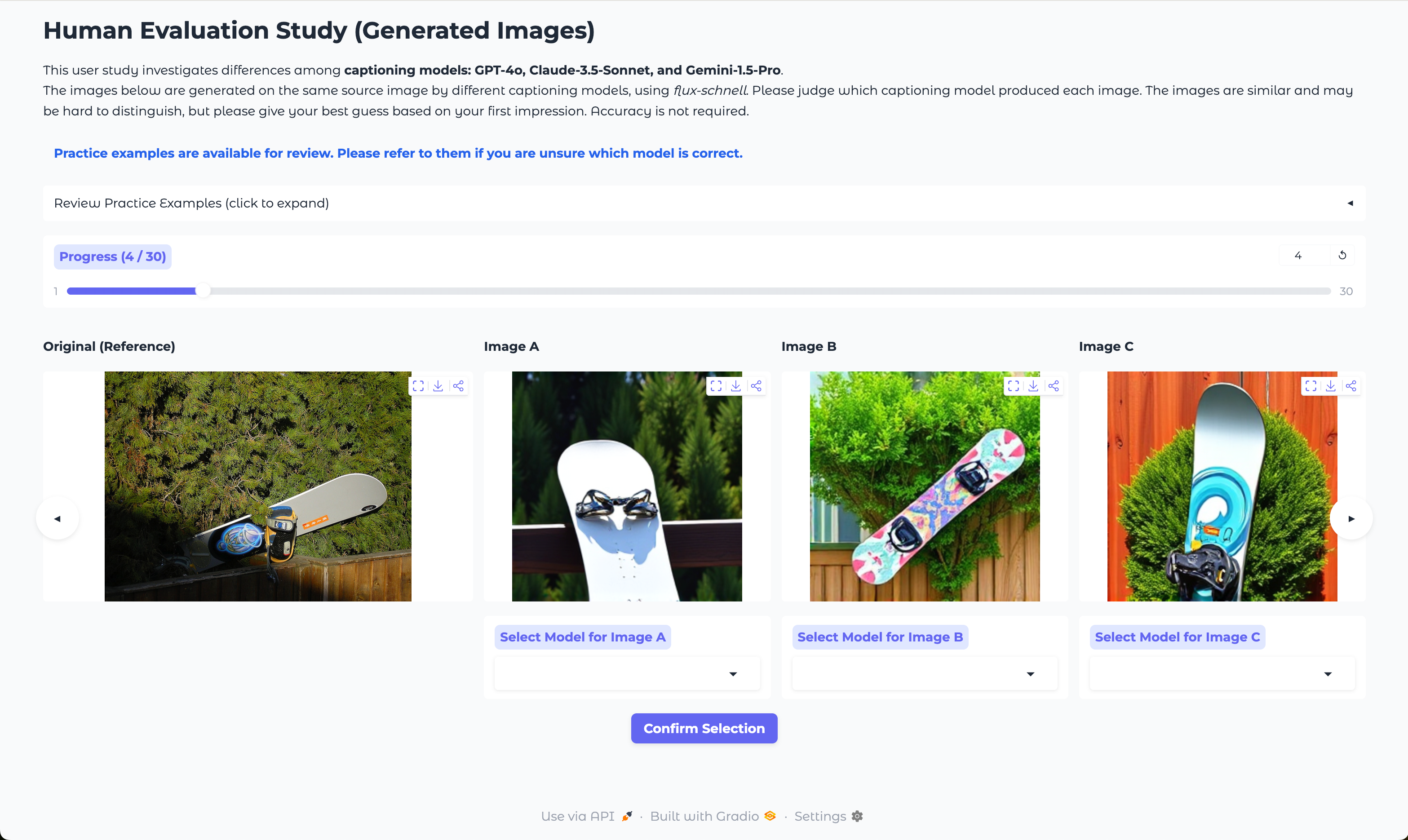}
    \vspace{6pt}
    \caption{User study interface for caption and image attribution. Left: caption. Right: image. Top: examples. Bottom: test.}
    \label{fig:user_study}
\end{figure*}

We conduct a small-scale user study to establish a human baseline for whether captioner-specific ``fingerprints'' are perceptible in (i) captions and (ii) the corresponding text-to-image (T2I) outputs (Fig.~\ref{fig:user_study}). For each task, participants first review 50 training examples with ground-truth context: the original image plus either three candidate captions (caption attribution) or three FLUX.1-generated images (image attribution). They then complete two independent tests on held-out prompts: 90 caption-only attribution trials and 90 image-only attribution trials on FLUX.1 outputs; for image attribution, we additionally display the original image to reduce ambiguity from prompt underspecification. We recruit 15 participants per task (10 with math or AI research backgrounds). Humans achieve \textbf{78.37\%} accuracy on caption attribution but only \textbf{41.63\%} on image attribution, suggesting that stylistic cues are evident in captions yet substantially attenuated after T2I generation.

\section{Paraphrase Prompts}
\label{app:paraphrase_prompts}

These are the three distinct prompts to generate paraphrased versions of captions, applied with both Qwen-2.5-1.5B and Qwen-2.5-7B to ensure that the robustness analysis is not tied to a single paraphrase style. 

\paragraph{Prompt 1.}
\begin{boxF}
\texttt{\{src text\}\\
Paraphrase:}
\end{boxF}


\paragraph{Prompt 2.}
\begin{boxF}
\texttt{Paraphrase the following text while maintaining the semantic meaning of the original text.\\
\{src text\}\\
Paraphrase:}
\end{boxF}


\paragraph{Prompt 3.}
\begin{boxF}
\texttt{Paraphrase the following text while maintaining the semantic meaning of the original text. Do not add explanations, suggestions, or follow-up questions. Only output the paraphrased text.\\
\{src text\}\\
Paraphrase:}
\end{boxF}

\section{LLM analysis on the characteristics of the generated captions}
\label{app:full_analysis}
This part contains the complete text from three different large language models tasked with analyzing the distinctive features of captions generated by Claude-3.5-Sonnet, Gemini-1.5-Pro, and GPT-4o. The text is presented as originally generated, with only minor typographical edits to remove non-standard characters (e.g., emojis) for display compatibility.

\subsection*{Analysis from ``Claude-Sonnet 4''}
\begin{boxF}
Looking at the captioning outputs from these three vision language models across 10,000 images, I can identify several distinctive language patterns for each:

\subsection*{Claude (claude-3-5-sonnet)}
\paragraph{Distinctive Language Features:}
\begin{itemize}
    \item Conversational and natural tone: Uses phrases like ``This image shows...'' or ``In this image, we can see...''
    \item Concise yet descriptive: Provides good detail without being overly verbose.
    \item Contextual inference: Often adds interpretive context (e.g., ``appears to be part of an organized event'').
    \item Practical descriptions: Focuses on what's practically happening in scenes.
    \item Moderate technical vocabulary: Uses specific terms when appropriate but remains accessible.
    \item Smooth narrative flow: Descriptions read naturally without feeling mechanical.
\end{itemize}
\paragraph{Identifying Markers:}
\begin{itemize}
    \item Opening phrases: ``This image shows...'', ``This is a...''
    \item Contextual additions: ``suggesting...'', ``appears to be...'', ``likely...''
    \item Balanced detail level - not too sparse, not exhaustively detailed.
    \item Natural sentence structure and flow.
\end{itemize}

\subsection*{Gemini (gemini-1.5-pro)}
\paragraph{Distinctive Language Features:}
\begin{itemize}
    \item Highly structured format: Consistently uses viewpoint descriptions (``Close-up, slightly low-angle view...'').
    \item Extremely detailed and systematic: Provides exhaustive catalogs of visible elements.
    \item Technical precision: Uses specific photography/cinematography terms.
    \item Organized presentation: Often uses bullet points or structured breakdowns.
    \item Spatial awareness: Detailed positioning and relationship descriptions.
    \item Comprehensive coverage: Attempts to describe virtually every visible element.
\end{itemize}
\paragraph{Identifying Markers:}
\begin{itemize}
    \item Signature opening: Almost always starts with viewing angle/perspective description.
    \item Systematic enumeration: Lists elements methodically.
    \item Technical terminology: ``Eye-level view'', ``low-angle'', ``close-up'', etc.
    \item Spatial precision: Detailed positional relationships.
    \item Exhaustive detail: Much longer descriptions than other models.
    \item Structured formatting: Uses line breaks and organized sections.
\end{itemize}

\subsection*{GPT-4o (gpt-4o)}
\paragraph{Distinctive Language Features:}
\begin{itemize}
    \item Balanced and measured: Moderate detail level between Claude's brevity and Gemini's exhaustiveness.
    \item Slightly more formal tone: Professional but not overly technical.
    \item Thematic focus: Often emphasizes the overall scene or atmosphere.
    \item Descriptive but efficient: Good coverage without excessive detail.
    \item Clear and straightforward: Direct descriptions without much interpretive context.
    \item Consistent structure: Reliable format across different image types.
\end{itemize}
\paragraph{Identifying Markers:}
\begin{itemize}
    \item Standard opening: ``The image shows...'' or ``The image depicts...''
    \item Moderate detail level: More detailed than Claude, less exhaustive than Gemini.
    \item Atmospheric descriptions: Often mentions mood, lighting, or overall impression.
    \item Balanced coverage: Describes main elements without getting lost in minutiae.
    \item Professional tone: Formal but accessible language.
    \item Consistent length: Descriptions tend to be similar in scope across images.
\end{itemize}

\paragraph{Key Distinguishing Patterns:}
\begin{itemize}
    \item \textbf{Length:} Gemini $>$ GPT-4o $>$ Claude
    \item \textbf{Technical detail:} Gemini $>$ GPT-4o $>$ Claude
    \item \textbf{Conversational tone:} Claude $>$ GPT-4o $>$ Gemini
    \item \textbf{Structural consistency:} Gemini $>$ GPT-4o $>$ Claude
    \item \textbf{Interpretive context:} Claude $>$ GPT-4o $>$ Gemini
\end{itemize}

\paragraph{Most Reliable Identifiers:}
\begin{itemize}
    \item \textbf{Gemini:} Look for viewing angle descriptions at the start and systematic, exhaustive detail.
    \item \textbf{Claude:} Look for conversational tone, contextual inference, and natural flow.
    \item \textbf{GPT-4o:} Look for ``The image shows...'' openings and balanced, professional descriptions.
\end{itemize}
\vspace{1em}
Based on my analysis of the captioning samples, each model has very distinct linguistic fingerprints that make them quite identifiable: Gemini is the easiest to spot - it almost invariably opens with precise viewpoint descriptions like ``Close-up, slightly low-angle view of...'' and provides extremely systematic, exhaustive detail. It reads like a professional photography or cinematography analysis. Claude has a conversational, natural tone with contextual interpretation. It tends to be concise but adds meaningful context about what might be happening or the purpose of scenes. GPT-4o sits in the middle ground with professional but accessible language, moderate detail levels, and tends to emphasize overall atmosphere or themes. The most reliable identifying feature is probably Gemini's signature opening format describing camera angles and viewing perspectives, which appears consistently across all its samples. Claude's interpretive context-adding (``appears to be,'' ``suggesting'') and GPT-4o's balanced, thematic approach are also quite distinctive.
\end{boxF}

\subsection*{Analysis from ``GPT-5''}
\begin{boxF}
Here’s a synthesis of the distinctive language identities of the three captioning models (Claude-3.5-Sonnet, Gemini-1.5-Pro, and GPT-4o), based on the multi-example outputs you provided. I’ll break this into model-by-model profiles, then give you practical “fingerprints” that can help you recognize each model in unseen captions.
\subsection*{Claude-3.5-Sonnet}
\paragraph{Stylistic Features:}
\begin{itemize}
    \item \textbf{Concise but polished:} Usually 1–2 medium paragraphs, not overly long. Sentences are smooth and descriptive, but not overloaded with details.
    \item \textbf{Interpretive layer:} Often adds contextual or interpretive remarks (e.g., ``suggests this is likely part of a European palace'', ``creating a humorous modern zombie moment'', ``perfect for a special occasion'').
    \item \textbf{Elegant phrasing:} Uses words like ``captured perfectly,'' ``haunting,'' ``tranquil,'' ``reminiscent of''. Feels almost like editorial writing or a photo-essay caption.
    \item \textbf{Specific but not hyper-granular:} Names obvious objects but rarely catalogs every tiny detail (compared to Gemini).
    \item \textbf{Balanced register:} Mixes factual description with mood/atmosphere, often noting lighting, ambience, or historical/cultural context.
\end{itemize}
\paragraph{Identities:}
\begin{itemize}
    \item Recognizable by interpretive commentary + smooth, compact prose.
    \item Frequent patterns: ``This image shows…'', ``The scene captures…'', ``The overall aesthetic suggests…''.
    \item Captions read like professional magazine blurbs, not technical listings.
\end{itemize}

\subsection*{Gemini-1.5-Pro}
\paragraph{Stylistic Features:}
\begin{itemize}
    \item \textbf{Extremely detailed \& structured:} Often breaks descriptions into bullet-style lists or sections (``Here’s a breakdown of the visible elements'').
    \item \textbf{Low-angle / high-angle qualifiers:} Very consistent in noting camera perspective (``eye-level view,'' ``slightly low-angle,'' ``close-up, over-the-shoulder'').
    \item \textbf{Meticulous object listing:} Names minutiae others ignore (e.g., ``a red tie holds scallions,'' ``a skull and crossbones forward of the name,'' ``a diamond-shaped pesticide sign'').
    \item \textbf{Color \& texture obsession:} Adds precise qualifiers like ``muted grey-blue,'' ``light brown potatoes,'' ``bubble canopy,'' ``murky brown water''.
    \item \textbf{Long, multi-paragraph outputs:} Typically 3–5 dense blocks, making it the most verbose of the three.
\end{itemize}
\paragraph{Identities:}
\begin{itemize}
    \item Look for explicit scene geometry, camera angles, exhaustive object breakdowns, fine-grained texture/color.
    \item Common intro: ``Eye-level view of…'', ``Close-up, slightly low-angle view…''.
    \item Tone feels technical + photographic, almost like a field notebook or art catalog.
\end{itemize}

\subsection*{GPT-4o}
\paragraph{Stylistic Features:}
\begin{itemize}
    \item \textbf{Compact and plain:} Shortest answers overall, usually one paragraph (sometimes two).
    \item \textbf{Straightforward object naming:} Lists major entities but skips minor details.
    \item \textbf{Mild contextualization:} Notes obvious atmospheres (``festive,'' ``tranquil,'' ``theatrical''), but less interpretive than Claude.
    \item \textbf{Generalized adjectives:} Uses words like ``lively,'' ``picturesque,'' ``ornate,'' ``beautiful'' without the subtle nuance Claude uses.
    \item \textbf{Dataset-savvy phrasing:} Occasionally references blurriness, motion, or resemblance to ML datasets (``reminiscent of MNIST digits,'' ``slightly blurry, indicating movement'').
\end{itemize}
\paragraph{Identities:}
\begin{itemize}
    \item Recognizable by brevity + general language.
    \item Captions feel functional, like ``answering the prompt efficiently'' rather than telling a story.
    \item Common phrasing: ``The image shows…'', ``The scene features…'', ``In the background…''.
\end{itemize}

\paragraph{Quick Fingerprints (How to Spot Them Fast):}
\begin{itemize}
    \item \textbf{Claude-3.5-Sonnet} $\rightarrow$ Interpretive + elegant, mid-length, adds atmosphere/context (sounds like a curated photo caption).
    \item \textbf{Gemini-1.5-Pro} $\rightarrow$ Verbose + hyper-detailed, mentions camera angle/lighting/textures, sometimes bullet-point structure.
    \item \textbf{GPT-4o} $\rightarrow$ Short + simple, names essentials, minimal elaboration, sometimes dataset-aware phrasing.
\end{itemize}

Would you like me to build a reference table with ``tell-tale phrases/structures'' (almost like linguistic fingerprints) so you can quickly identify which model likely produced a caption just by scanning it? That could help if you’re planning classification or human evaluation.
\end{boxF}

\subsection*{Analysis from ``Gemini-2.5 Pro''}
\begin{boxF}
Based on the provided captioning outputs, each vision-language model demonstrates a distinct and recognizable linguistic style.

\subsection*{Claude 3.5 Sonnet: The Narrative Storyteller}
Claude's captions are characterized by their interpretive and narrative quality. It often goes beyond a simple description to infer the context, mood, or story behind the image. Its language is frequently evocative and focused on the overall atmosphere.
\paragraph{Distinctive Features \& Identities:}
\begin{itemize}
    \item \textbf{Atmospheric Language:} Uses descriptive adjectives that convey a feeling or mood, such as ``dynamic quality,'' ``intimate perspective,'' ``magical atmosphere,'' or ``mysterious and haunting appearance.''
    \item \textbf{Contextual Inference:} Frequently makes educated guesses about the event or location, such as ``a zombie walk or Halloween event,'' ``a dog show,'' or ``a classic Seattle holiday scene.''
    \item \textbf{Narrative Flow:} Presents descriptions in a single, flowing prose paragraph that reads like a short story or a scene description.
    \item \textbf{Potential Weakness:} Can sometimes misinterpret abstract or non-standard images by trying to force a real-world context. For example, it incorrectly identified a handwritten digit from the MNIST dataset as a ``basketball jersey.''
\end{itemize}

\subsection*{Gemini 1.5 Pro: The Methodical Analyst}
Gemini's style is highly structured, detailed, and analytical. Its primary goal appears to be an exhaustive and objective cataloging of the visual information, often organized in a systematic way that is easy to parse.
\paragraph{Distinctive Features \& Identities:}
\begin{itemize}
    \item \textbf{Camera Perspective Opening:} A very strong identifier is that its captions almost always begin by stating the camera's angle and proximity, such as ``Close-up, slightly low-angle view,'' ``Eye-level view,'' or ``Over-the-shoulder view.''
    \item \textbf{Structured Formatting:} Frequently uses double newlines to separate paragraphs, creating distinct logical blocks. For detailed prompts, it may even use bulleted lists with bolded headings to break down every component in the image.
    \item \textbf{Exhaustive Detail:} Tends to produce the longest and most granular descriptions, mentioning specific details like skin tone, indistinct graphics on a shirt, or the material of an object (e.g., ``matte finish'' on scissor handles).
    \item \textbf{Objective Tone:} The language is typically encyclopedic and neutral, focusing on what is visually present rather than the emotional tone of the scene.
\end{itemize}

\subsection*{GPT-4o: The Efficient Summarizer}
GPT-4o is defined by its conciseness and directness. It excels at providing a quick, accurate, and highly efficient summary of the image's content without unnecessary elaboration or stylistic flair.
\paragraph{Distinctive Features \& Identities:}
\begin{itemize}
    \item \textbf{Brevity:} Its captions are consistently the shortest of the three models. It distills the scene to its most essential elements.
    \item \textbf{Direct-to-the-Point:} It avoids introductory framing (like Gemini's camera angles) and immediately states the main subject of the image.
    \item \textbf{Balanced Tone:} The tone is generally neutral and factual but can incorporate a touch of descriptive language (e.g., ``tranquil and picturesque atmosphere'') when it adds significant value.
    \item \textbf{High-Level Accuracy:} It demonstrates strong and accurate recognition, even with abstract subjects. For instance, it not only identified the handwritten digit correctly but also astutely noted its resemblance to the ``MNIST dataset,'' showing deep contextual knowledge.
\end{itemize}
\end{boxF}

\section{Use of Large Language Models}
Large language models were used in this work to assist with writing and editing tasks, such as polishing grammar, improving clarity, and suggesting alternative phrasings for sections of the manuscript. No LLM outputs were used as scientific claims, experimental findings, or conclusions. The authors take full responsibility for the accuracy and integrity of all content presented in this paper.


\end{document}